# Bringing Back Rule Induction to Fluid Intelligence Research? An Initial Validation of the ARC-AGI Benchmark in Humans

Jasmin Thelen[1], Oliver Wilhelm[1]

[1] Institute of Psychology and Education, Ulm University, Germany

## Author Note

Jasmin Thelen https://orcid.org/0000-0002-3468-6894

Oliver Wilhelm https://orcid.org/0000-0001-7980-1166

We did not receive any financial support for the research, authorship, and/or publication of this article. We confirm that the work conforms to Standard 8 of the American Psychological Association's Ethical Principles of Psychologists and Code of Conduct. All data and code are available at https://osf.io/b2ugc.

Correspondence concerning this article should be addressed to Jasmin Thelen, Institute of Psychology and Education, Ulm University, Albert-Einstein-Allee 47, 89081 Ulm, Germany, Email: jasmin.thelen@uni-ulm.de

## Abstract

Two competing perspectives on fluid intelligence (gf) measures propose that performance is primarily constrained either by working memory capacity or by the ability to induce novel relations. The first perspective is currently dominant in measurement, as evident from the use of a limited set of recurring rules, whereas the second perspective is reflected in many definitions but rarely present in measurement. The ARC-AGI benchmark predominantly requires rule induction and was proposed as a measure of gf for both humans and artificial systems. However, its psychometric properties have not yet been examined in human samples. We therefore investigated the psychometric characteristics and nomological network of ARC-AGI in a first study with 100 participants. A compilation of ARC-AGI items showed good psychometric properties and correlated substantially with figural fluid intelligence as measured by a figural reasoning test ($\rho = .63$). Associations with figural originality were weak. These findings provide initial support for the validity of ARC-AGI as a measure of human fluid intelligence. Future research should include more rule induction tasks as well as additional multivariate covariates. This study is unusual by studying a task in humans that was initially designed for machines. We suggest systematically embedding AI benchmarks into the nomological network of human cognitive abilities to enable more systematic evaluation and interdisciplinary cooperation.

Words: 213

# Bringing Back Rule Induction to Fluid Intelligence Research? An Initial Validation of the ARC-AGI Benchmark in Humans

Fluid intelligence (gf) is an indispensable and prototypical part of human cognitive abilities. It is the cognitive ability most strongly associated with general intelligence (Carroll, 1993) and consistently predicts academic achievement, job performance, and health outcomes (Gottfredson & Deary, 2004; Kuncel et al., 2001, 2004; Sackett et al., 2022; Watrin et al., 2022). Many definitions emphasize novel information processing as essential for gf (W. J. Schneider & McGrew, 2018), yet established gf tests typically rely on a relatively small set of rules that are repeatedly used across items (e.g. Becker et al., 2016; Carpenter et al., 1990; Schroeders & Walter, 2026). This raises the question whether current measures include novel information processing to a sufficient degree.

Recently, gf gained considerable relevance outside of psychology: it has been incorporated into the evaluation of artificial intelligence (AI) (Hendrycks et al., 2025; Qu et al., 2024; Song et al., 2024). Presumably the most prominent example is the Abstraction and Reasoning Corpus for Artificial General Intelligence (ARC-AGI, Chollet, 2019; Chollet et al., 2025). It is designed as a gf measure for both humans and machines, with an explicit focus on rule finding. For years, AI models struggled to achieve meaningful performance on ARC-AGI (Chollet et al., 2025), motivating competitions and research (ARC Prize, 2026a; Chollet et al., 2025, 2026). Yet, whether ARC-AGI measures fluid intelligence in humans has remained an empirically untested claim.

This gap matters for two reasons. First, the emphasis on novel rules of ARC-AGI could enrich human intelligence measurement, given that popular gf tasks do not stress novel information processing notably. Second, without establishing ARC-AGI as a valid measure of human gf, it remains unclear, whether AI performance on the benchmark reflects anything meaningfully comparable to human intelligence. In the present study, we therefore examine how ARC-AGI correlates with established measures of fluid intelligence in human samples. In doing so, we seek to demonstrate that new and innovative AI task designs can enrich the measures of human cognitive abilities, and that validation of AI benchmarks in human samples can strengthen AI evaluation.

**Definition of Fluid Intelligence**

Fluid intelligence (gf) is the most central and prototypical cognitive ability within many intelligence models (Carroll, 1993; Cattell, 1987; W. J. Schneider & McGrew, 2018). Gf is characterized by "understanding relations among stimuli, comprehend implications, and draw inferences" (Horn & Noll, 1997, p. 69). Two perspectives on gf have emerged: one emphasizing novelty, the other domain-general cognitive capacity. The following sections outline both and highlight key outstanding issues.

**Fluid Intelligence as Novel Information Processing**

Solving novel situations is central to both scientific definitions of fluid intelligence (Gignac & Szodorai, 2024; W. J. Schneider & McGrew, 2018) and lay conceptions (Berg & Sternberg, 1992). Sternberg (1981, 1984) proposed that intelligence is most clearly demonstrated when adapting to novelty. However, novelty itself is rarely defined. Spearman (1923) cautioned that defining intelligence with novelty means explaining one obscure construct with an even more obscure one. The measurement of novelty is also difficult. Sternberg and Gastel (1989) for example, assessed the ability to deal with novelty by giving participants a premise that was either true in our world (e.g., "Cats are furry") or true only in a fictional world (e.g., "Cats are strongly magnetized"). Participants then judged a statement based on the premise (e.g., "Cats stick to refrigerators," true only under the fictional premise). Better performance on this task was related to higher intelligence test scores. However, Larson (1990) noted that the novel condition inadvertently increased working memory load because participants had to maintain additional information, which confounded the results.

Tasks designed to elicit novel information processing, such as insight problems (Sternberg, 1984), are strongly related with standard gf tasks (Chuderski & Jastrzębski, 2018) and arguably they capture little beyond that. Thus, both the definition and the measurement of novelty appears to be an unsolved issue and its measure has not been established in gf tasks. Nevertheless, it is still stressed in many popular definitions of gf.

Among gf's subcomponents, one could argue that induction aims for identifying novel relations. (Carroll, 1993; W. J. Schneider & McGrew, 2018). It refers to the "ability to observe a phenomenon and discover the underlying principles or rules that determine its behavior" (W. J. Schneider & McGrew, 2018, p. 93). The primary task is to *identify* underlying rules (Shye, 1988).

Popular induction tasks are best considered to be enthymemes, because the required inferences do not imply an increase in semantic content beyond what is explicitly or implicitly provided in the premises (Wilhelm, 2005). A common example is number series tasks, in which the instructions do not explicitly specify which mathematical operations will be used. Nevertheless, the use of the four basic arithmetic operations (addition, subtraction, multiplication, and division) can be implicitly assumed by subjects. Solvers are therefore not required to generate entirely novel rules but rather to identify which of a small set of conventionally available operations best accounts for the presented sequence.

Another approach to measure induction is concept formation (Horn, 1976; Wilhelm & Kyllonen, 2021; Woodcock, 1990). Rooted in experimental psychology, it refers to grouping events, objects, or ideas into abstract classes (i.e. concepts) based on shared properties (Bourne, 1966; Seel, 2012). Successful classification requires identifying commonalities across different instances (Smoke, 1932), from which rules can be derived to distinguish members of a concept from non-members (Hunt, 1961). Although mentioned in Carroll's (1993) analysis of ability tasks, to the best of our knowledge, in contemporary intelligence tests, concept formation is presumably measured only in the Woodcock-Johnson Tests of Cognitive Abilities. In the Concept Formation subtest, participants are shown several simple geometric shapes that vary in size, color, and form. The shapes form two groups: Some shapes are each outlined by a frame, and the remaining forms have no frame. Participants have to name the rule that determines when shapes are framed (Schrank et al., 2014).

Many figural matrices tests also require rule identification (Amthauer et al., 1999; Raven et al., 1998; Weiß, 2006) and are therefore classified as inductive (Wilhelm, 2005). We argue, however, that rule identification is not central here. Most established tests rely on a small set of recurring rules (e.g., Becker et al., 2016; Carpenter et al., 1990; Schroeders & Walter, 2026). When providing participants with explanations of the underlying rules prior to testing in one condition, performance improved substantially (Loesche et al., 2015; B. Schneider et al., 2020). However, no ceiling effect was present, showing that the tasks without inducing new rules were not trivial. Metric invariance of gf factor loadings was given across conditions, meaning rule provision did not alter items' relevance to the latent factor (Levacher et al., 2022; Loesche et al., 2015). Empirically, induction tests do not form a separable factor from gf. Instead, fluid intelligence is best understood as a higher-order factor above figural/spatial, numerical, and verbal first-order factors (Wilhelm, 2005).

In sum, popular inductive tasks continue to function equally well when participants are provided with the underlying rules and other tasks focusing on novelty are rarely used. This is leading to the second perspective on fluid intelligence.

**Fluid Intelligence as Working Memory**

Another prominent approach frames fluid intelligence as being largely determined by working memory capacity (WMC), the capacity to maintain and update mental representations (Kyllonen & Christal, 1990). This view is supported by the strong empirical overlap of gf and WMC of $\rho$ = .70 - .90 (Kane et al., 2005; Kyllonen & Christal, 1990;

Monteiro et al., 2025; Oberauer et al., 2005; Wilhelm et al., 2013). In typical WMC tasks, both the problem format and the solution procedure are explicitly specified in advance. Complexity in gf tasks results from increasing working memory load (Carpenter et al., 1990) meaning that as the number of elements increases, so does item difficulty. Empirically, the number of rules or solution steps is a strong predictor of difficulty, whereas the specific type of rule tends to have limited effects (Carpenter et al., 1990; Cortes et al., 2021; Gabales & Birney, 2011; Hartung et al., 2022; Primi, 2001).

In sum, performance in prevalent gf measures is largely limited by working memory. Gf tasks show strong associations with working memory tasks, that have no demand of novel information processing. When controlling for WMC, only a small proportion of variance in gf remains unexplained. Thus, either definitions of gf overemphasize novel information processing, or current gf tasks are incompletely populating the construct space of gf. Then measures that challenge rule induction are desperately needed.

**Novel Information Processing in the Abstraction and Reasoning Corpus for Artificial General Intelligence (ARC-AGI)**

Inspiration for rule-identifying tasks comes from computer science. ARC-AGI (Chollet, 2019) explicitly emphasizes the discovery of diverse underlying rules in a large and diverse item set and thus may better capture aspects of fluid intelligence underrepresented in current measures. ARC-AGI is designed to measure fluid intelligence in humans and machines (Chollet, 2019). For each task, an underlying rule must be derived from a set of input-output pairs and then applied by generating the correct output grid for a new input (see Figure 1). The difference to many established gf tasks is that most task contain a new rule to infer. In ARC-AGI-1, 400 public training tasks and 400 public test tasks are available (Chollet, 2019).

**Figure 1**

*Example Task of ARC-AGI 1*

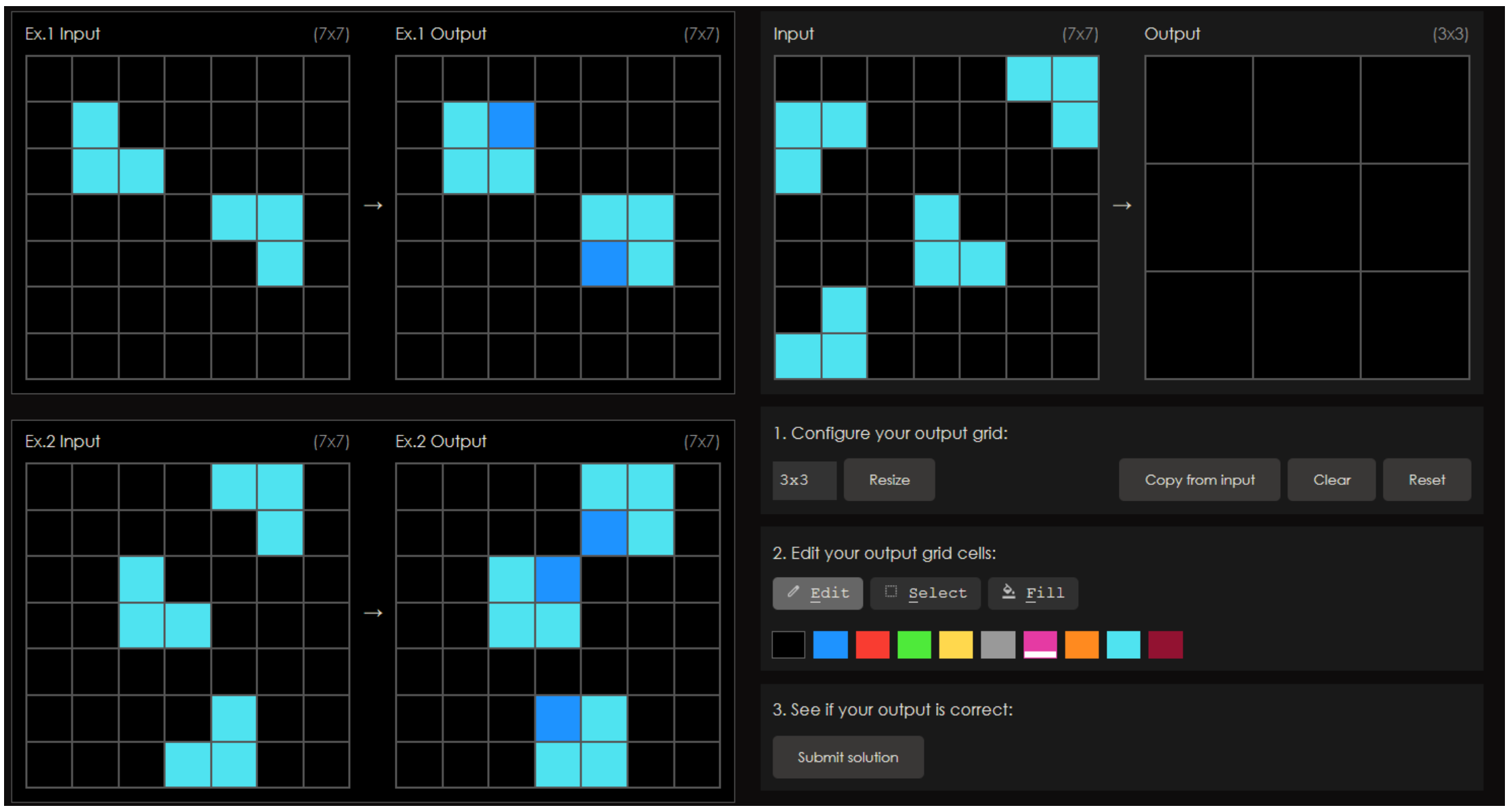

*Note*. The left panel shows two example input - output pairs. The participant's task is to infer the transformation rule that maps an input to its corresponding output across all examples. They must construct the correct output in the right panel. In this example, the underlying rule is to complete each turquoise L-shaped figure into a full 2×2 square by adding a blue cell at the missing corner position. The example corresponds to task 3aa6fb7a from the public training dataset of ARC Prize ARC-AGI-1, retrieved from ARC Prize (2026c).

*Rule Diversity in ARC-AGI*

Building on the priors outlined by Chollet (2019), we classified the underlying rules represented in the evaluation set. This classification resulted in 15 rule categories. To illustrate the broadness of the rule categories, we discuss two items of the same rule category in more detail (see Figure 2). Both items require selecting a subpart of the input grid to construct the output, which provides the basis for their classification within this category. However, the specific underlying rule differs considerably. In Panel A, the task requires identifying the rectangle that contains the smallest number of yellow and pink pixels and reproducing that rectangle as the output. In Panel B, the task requires identifying all red pixels are horizontally surrounded by a dark blue pixel to their left and a light blue pixel to their right. Only these red pixels are used in the output.

**Figure 2**

*Illustration of Rule Diversity within ARC-AGI 1*

A)

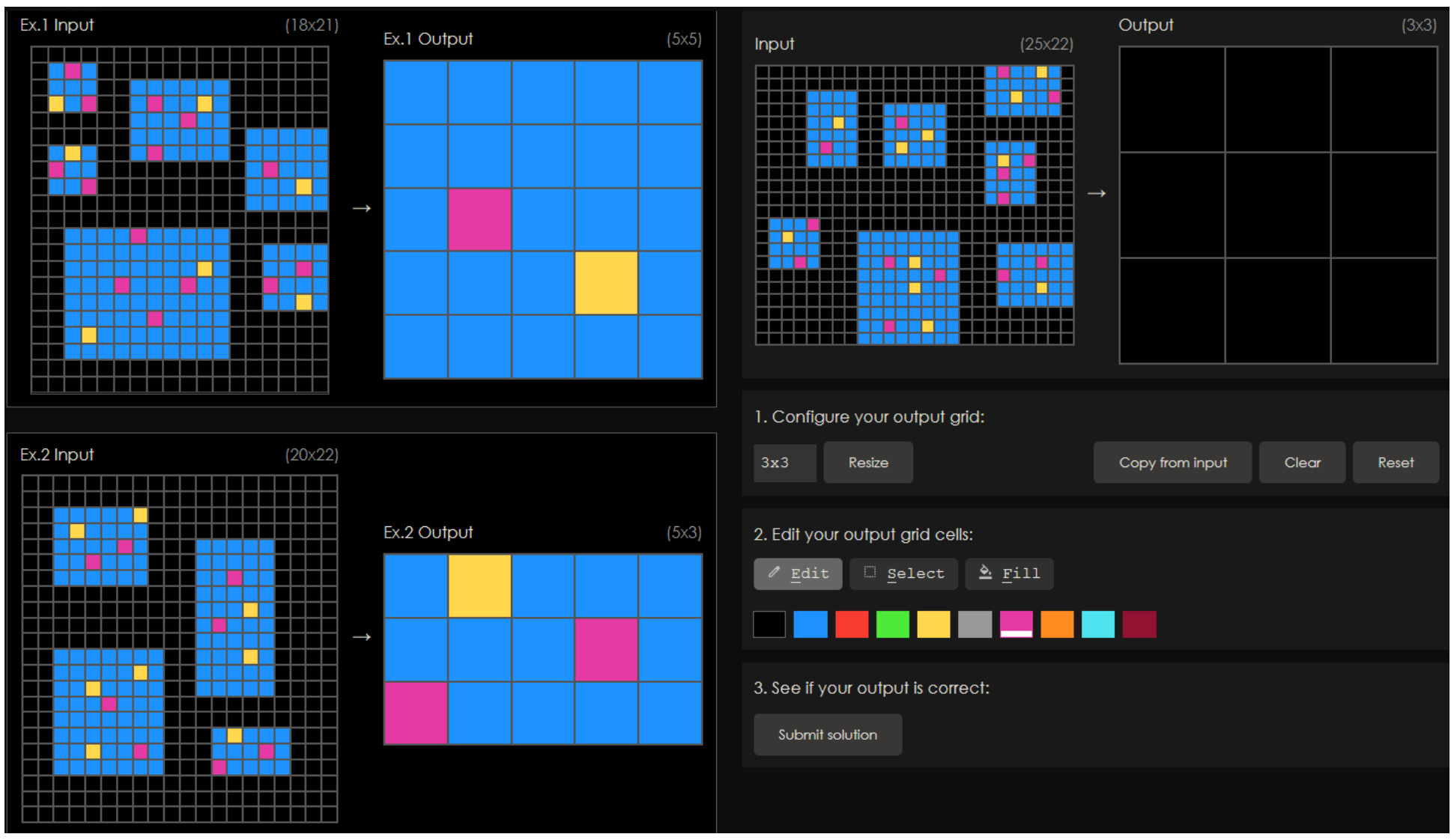

B)

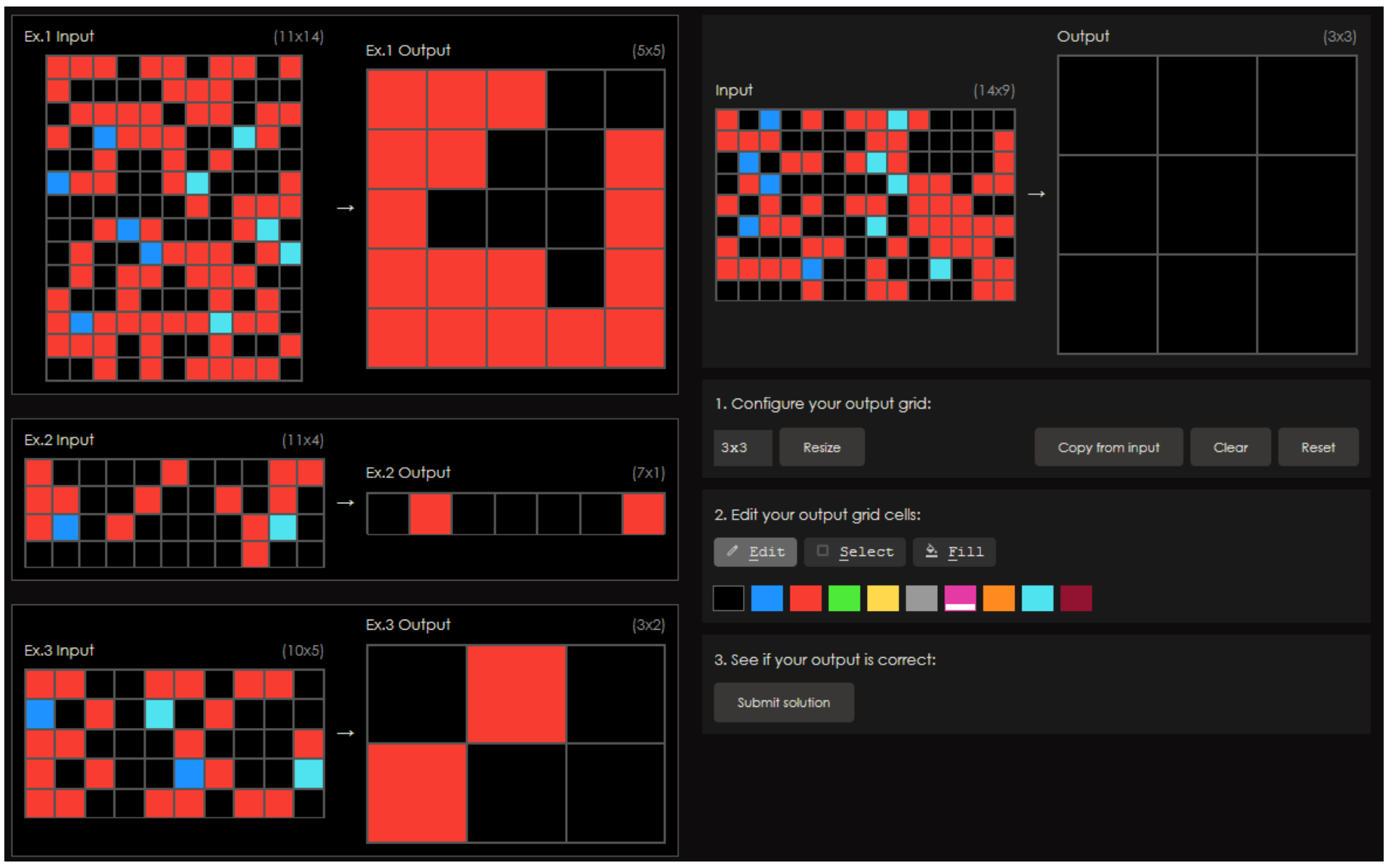

*Note.* The examples correspond to A) task 7bb29440 and B) task 505fff84 from the public evaluation dataset of ARC Prize ARC-AGI-1, retrieved from ARC Prize (2026c). Only the first Input-Output examples of the tasks are shown for the sake of simplicity.

## *ARC-AGI in Humans*

To evaluate ARC-AGI as a gf measure for humans, several psychometric criteria must be met. Tasks must be (1) solvable by humans (2) sufficiently varied in difficulty to capture meaningful variance across the full ability range (Cohen, 2010; Loevinger, 1947) and (3) a measurement model must be established (Borsboom et al., 2004). If gf is a limiting factor for various ARC-AGI items, their intercorrelations should vanish once a single latent factor is controlled for. (4) The latent factor of ARC-AGI should correlate with established fluid intelligence tasks to demonstrate convergent validity (Cronbach & Meehl, 1955). (5) ARC-AGI should be checked for its empirical link to other related constructs (nomological net, see Cronbach & Meehl, 1955).

The few existing human studies on ARC-AGI have focused on mean performance. Across three studies, average human accuracy with three attempts ranged from 65.7 % to 83.3 % (ARC Prize, 2026b; Johnson et al., 2021; LeGris et al., 2025) and dropped to 49.2% with a single attempt in the evaluation set (LeGris et al., 2025). Sample sizes were generally small, with a mean of 11.8 (training set, LeGris et al., 2025), 10.3 (evaluation set, LeGris et al., 2025) and 23.5 (Johnson et al., 2021) participants per task. ARC-AGI items can be solved by humans but inter-individual differences have not been studied so far and thus the measurement model as well as the correlation with fluid intelligence and other covariates have not yet been investigated. Next, we will derive different perspectives on the nomological net of ARC-AGI, building upon the perspectives on fluid intelligence mentioned above.

**The Nomological Net of ARC-AGI**

Under the view that novel information processing is a central component of fluid intelligence, a positive correlation between ARC-AGI performance and other measures of fluid intelligence is expected. In contemporary models of intelligence, inductive reasoning is conceptualized as a subcomponent of fluid intelligence (W. J. Schneider & McGrew, 2018; Wilhelm & Kyllonen, 2021). As noted above, ARC-AGI tasks involve a large number of rules that must be inferred, thereby fulfilling key requirements for measuring inductive reasoning. However, because many established measures of fluid intelligence do not strongly emphasize the inductive component, moderate correlations between ARC-AGI performance and these traditional measures would be expected in this perspective.

When aiming to enhance induction in the measurement of gf, divergent thinking (DT) emerges as a promising covariate. DT is characterised by the generation of multiple ideas to address a specific problem (Guilford, 1967) and is considered a marker of creative potential (Runco & Acar, 2012) and creative achievement (Kim, 2008; Runco et al., 2010). Among the established scoring methods for DT tasks, originality captures the quality of creative ideas and thus expresses unusual but still appropriate responses (Carroll, 1993; Guilford, 1956; Weiss et al., 2024). Following this, originality may facilitate inductive rule discovery. Subjects who generate unusual yet valid mappings may hold an advantage in ARC-AGI performance. This expectation is supported by a meta-analytic manifest correlation of gf and DT of $r = .25$, whereby the effect was equally high for g. The correlation of g and DT increased to a manifest correlation of $r = .34$ when subjects were instructed to be creative and scored for originality (Gerwig et al., 2021).

From the perspective that fluid intelligence, including inductive reasoning, is primarily constrained by working memory capacity, the cognitive demands of rule induction may arise not from the induction process itself, but from the number of rules and rule candidates that must be simultaneously maintained in working memory. Solving ARC-AGI tasks requires individuals to concurrently maintain multiple candidate rules, compare them against available evidence, and continuously update these representations as additional constraints are detected. Accordingly, ARC-AGI performance and established measures of fluid intelligence are expected to show moderate positive associations with one another, while exhibiting only weak relationships with originality.

**Corollaries of Psychometric Principles for AGI Models**

Validating ARC-AGI with human subjects is not only interesting for psychology, it is also relevant for AI evaluation. ARC-AGI is a prominent example of numerous AI-benchmarks designed to evaluate performance of Large Language Models (Anthropic, 2025; OpenAI, 2026; Reuel et al., 2024). Benchmarks are standardized tests with predefined tasks, and defined correct answers. Like ARC-AGI, they typically consist of a large set of items (often 100 or more) that an AI model is supposed to solve. The mean proportion of correct answers is usually reported and compared across models in so-called leaderboards, that is, tables listing each model's name and its mean performance on the same items.

The quality of AI benchmarks has faced increasing criticism (Chizhov et al., 2025; see also Eriksson et al., 2025; Raji et al., 2021 for overviews). A central concern is limited generalizability. Models achieving high scores struggle when tasks are slightly reformulated, or answer options were reordered (Alzahrani et al., 2024; Pezeshkpour & Hruschka, 2023). In addition, what benchmarks intend to measure is often poorly defined (Bean et al., 2025; Blodgett et al., 2021). More broadly, benchmarks have been argued to fail to measure what they suggest to measure - or, in psychometric terms, their validity is limited (Alaa et al., 2025; Bean et al., 2025; Binette & Reiter, 2024; Eriksson et al., 2025). Next, we present the view of benchmarks as ability measures and show how the announced validation in humans can help to solve current issues with benchmarking in AI.

**AI Benchmarks as Ability Measures**

Ideally, an AI benchmark reflects a model's capability to solve a specific class of tasks, with the performance expected to generalize to similar problems. For instance, selecting a model based on strong performance on ARC-AGI would not be meaningful without the expectation that this performance translates to related reasoning tasks. Since testing all possible task variations is impossible, conclusions about general capabilities must be inferred from a limited set of tasks, meaning the underlying capability is not directly observable. This is analogous to psychology, where most phenomena, such as fluid intelligence, are likewise not directly observable. They are conceptualized as latent abilities, defined as “a postulated attribute of people, assumed to be reflected in test performance” (Cronbach & Meehl, 1955, p. 178). Crucially, different measures of the same ability are expected to correlate positively (convergent validity). Thus, benchmark performance is often implicitly treated as reflecting underlying abilities, even when not stated or tested by benchmark authors (Raji et al., 2021) and can be seen as terminologically correspond to psychological tests of latent constructs in psychometric research.

Accordingly, it is appropriate to accept the principle of indifferent indicators and the causal nature of unobservable abilities. Moreover, considering additional widely used psychometric validity criteria (Bean et al., 2025; Binette & Reiter, 2024; Reuel et al., 2024; Salaudeen et al., 2025; Wallach et al., 2025) seems pretty plausible and straight forward. Evidence points to positive correlation amongst ability-related benchmarks (Ilić & Gignac, 2024; Leelawat & Griffin, 2026; Song et al., 2024) mimicking what is known as positive manifold in human intelligence research.

The validation of AI benchmarks presents an additional challenge that differs from those encountered in the validation of traditional human ability tests. If AI benchmarks are intended to capture human-like intelligence, they “need to make sense within the nomological networks established for humans” (Sühr et al., 2025, p. 8). Consequently, the validation of a new benchmark should first demonstrate convergent validity with the corresponding human ability construct. For ARC-AGI, this means establishing a meaningful relationship with measures of fluid intelligence in human samples before claiming that the benchmark assesses fluid intelligence in AI models. This requirement is consistent with the definition of AGI (artificial general intelligence) as matching or exceeding human cognitive abilities, with humans described as the "only existing example of general intelligence" (Hendrycks et al., 2025, p. 2). If AGI is conceptualized in relation to human intelligence, construct validation of AI benchmarks should therefore include examining how benchmark performance relates to established human ability measures.

To the best of our knowledge, the evaluation of construct validity of AI benchmarks in relation to human abilities is still pending. Some studies have had language models complete human intelligence tests (Galatzer-Levy et al., 2024; Zhang & Wang, 2024), but this approach might be deemed problematic. Test items may have been present in training data (i.e. data leakage, see Hagendorff et al., 2024), meaning correct responses reflect memorized solutions rather than reasoning ability to an unknown degree. Validating benchmarks with human participants is therefore a critical and largely neglected step. A benchmark that fails to show convergent validity with established human intelligence measures is unlikely to meaningfully assess intelligence in artificial systems. ARC-AGI is supposedly well suited for a validation of benchmark tasks in humans. Humans can solve its tasks; it is widely used in AI evaluation, and its original paper called for a validation "via large-sample size statistical studies on humans"(Chollet, 2019, p. 54).

## The Current Study

ARC-AGI supposedly measures fluid intelligence in humans and machines. To date, no study has systematically examined its validity in humans. This gap is notable from a psychological perspective, given that ARC-AGI emphasizes the identification of underlying rules rather than their application, setting it apart from many frequently used gf tasks. From a computer science perspective, a validation of ARC-AGI within humans can contribute addressing the validity problem of benchmarks. In the current lab-study, $N = 100$ participants solved 20 carefully chosen ARC-AGI items from the evaluation set as well as an established gf test, a measure of concept formation, and figural originality. The aim is twofold: First, we aim to provide a psychometric evaluation of ARC-AGI in humans, by establishing a measurement model. Second, we examine the associations of ARC-AGI with figural originality and established measures of fluid intelligence. The results are discussed in the context of two competing theoretical perspectives: one that views inductive rule discovery as completion of fluid intelligence and another that emphasizes the shared influence of working memory capacity.

# Method

## Participants and Procedure

Participants were recruited through flyers, mailing lists, social media, and by advertisements in university courses. They received course credit for students or monetary compensation for their participation (15 €) and were given feedback on their task performance at the end of the study. Requirements for participation were no color vision deficiency and German language skills sufficient to understand the task instructions. Participants were required to be 18 years or older.

A local ethics committee decided that ethical approval for the study was dispensable. The study was conducted according to the declaration of Helsinki. Prior to the start of the study, all participants provided digital informed consent. Participants completed a computerized laboratory study in groups of up to six per session, under the supervision of a trained proctor. All tasks and instructions were presented on 24″ screens. The tasks were presented in the following, fixed order: Demographics, figural BEFKI, Set, MTCI and ARC-AGI (see below). The total test time was approximately 1.5 hours. The first three tests were administered using *Inquisit 6* (*Inquisit 6, Version 6.6.3*, 2022), MTCI was administered using *The Creativity Assessment Platform* (CAP, Patterson & Pronchick, 2024) and ARC-AGI was conducted through a modified version of the interface provided by Chollet (2019) via the R Shiny Package (*Version 1.10.0*, Chang et al., 2024).

105 subjects took part in the study. Participants who failed all three ARC-AGI example items ($n$ = 5; 5%) were excluded from all further analysis, as this suggested a lack of task comprehension. Additionally, we examined the data for multivariate outliers but did not identify any. Thus, the sample consisted of $N$ = 100 participants. 63% of the sample were female and 37% male. The average age was 22.4 years ($SD$ = 3.7, range = 18 - 44) and most participants were students (92%). 75% of respondents reported having completed high school and 24% reported having completed college. The majority of participants had German as their mother tongue (92%).

## Measures

For all tasks, participants submitted one attempt per item and did not receive feedback whether their answer was correct. If not stated differently, items were presented in ascending order of difficulty, beginning with the easiest.

### *Fluid Figural Intelligence*

Fluid figural intelligence was measured using the figural subtest of the *Berlin Test of Fluid and Crystallized Intelligence (BEFKI)* measure for persons of grade eleven and above in the upper school track within the German school system (Schipolowski et al., 2020). The test comprises a sequence of figures that systematically vary in position, shading, and rotation according to underlying rules. For an example task see Figure 2. Participants were required to complete each sequence by selecting the next two correct elements from three response options per position. An item was scored as correct if both selected elements were accurate. Else responses were coded as incorrect. The test included 16 items and was administered within a 15-minute time limit.

***Concept Formation***

A slightly modified version of the Set card game (Davis & Maclagan, 2003) was used in this study to measure concept formation. The task requires participants to identify groups of three cards, called "Sets", based on specific rules. Each card features symbols that vary along four dimensions: shape (cross, circle, square), color (blue, orange, purple), shading (solid, striped, open), and number of symbols (one, two, or three). A Set consists of three cards in which all but one feature were either completely identical or entirely different across all three selected cards. Participants were presented with either 9 or 12 cards, whereby exactly one unique combination of three cards formed a valid Set in each trial. For an example task see Figure 2. If all three correct cards were selected by participants, the item was scored as correct. Participants completed 8 items within a total time limit of 8 minutes.

***Figural Originality***

The Multi-Trial Creative Ideation (MTCI) task was used (Barbot, 2018) to measure figural originality. Participants were instructed to complete presented shapes in the most creative way possible, ensuring that the final drawing incorporated the original stimulus. For an example task see Figure 2. All six MTCI items were used and were presented in random order. Individuals had two minutes time per image.

**Figure 3**

*Example item from BEFKI figural, Set and MTCI.*

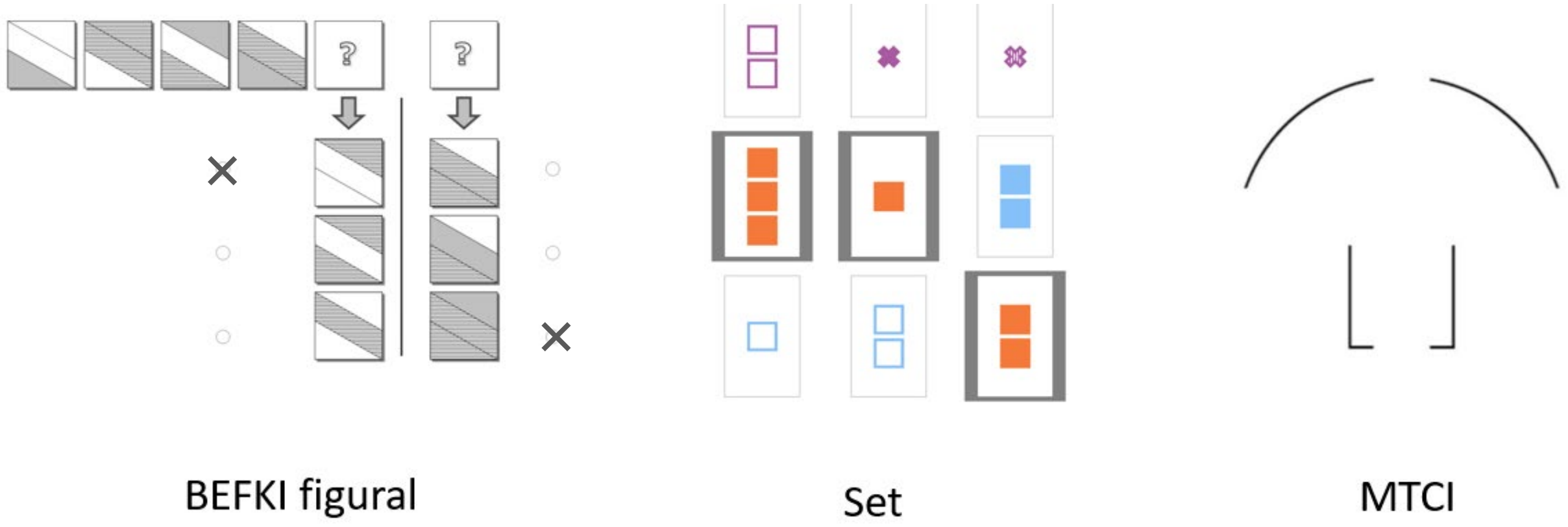

*Note*. figural BEFKI = figural subtest of Berlin Test of Fluid and Crystallized Intelligence, MTCI = Multi-Trial Creative Ideation task; for examples of ARC-AGI see Figure 1 and 2.

***The Abstraction and Reasoning Corpus for Artificial General Intelligence***

In the Abstraction and Reasoning Corpus (ARC-AGI, Chollet, 2019) each task presents a set of input-output grid pairs, where participants must identify the underlying transformation rule that connects them. Participants then apply this rule to a new input grid, determining the correct dimensions (width and height) of the output and colouring each pixel accordingly to the underlying rules (see Figure 1).

A subset of ARC-AGI 1 evaluation set was used. We decided for the evaluation item set, as first difficulty estimates showed that is more difficult than the training set (LeGris et al., 2025). We carefully inspected the 400-item evaluation set and used two human coders to rate the underlying rules of each item. The inter-rater reliability was good with cohen's kappa $\kappa = .75$ (Landis & Koch, 1977). A subset of 59 items, representing a diverse range of rules, was piloted with 24 to 40 psychology students per item (see appendix). Based on the pilot data, we selected items that covered a broad range of item difficulties while ensuring a diverse mix of rules. The final set comprised 20 items, with a time limit of 40 minutes for completion.

## Statistical Analysis

Items of figural BEFKI, Set and ARC-AGI were scored dichotomously with one if the participant provided the correct answer and zero if not. If an item was not completed due to lack of time, it was scored as false. Drawings were scored for originality by the AI model AuDrA, on a scale from 0 (low originality) to 1 (high originality). AuDrA was trained to reproduce human originality ratings of drawings from the MTCI task (Patterson et al., 2023). Because AuDrA can be accessed via CAP, all drawings were scored directly in CAP. Manifest correlations were tested against zero with one-sided tests, as positive relations were expected for all tasks.

All analysis were computed in R (Version 4.4.2 R Core Team, 2024), mostly relying on the packages *tidyverse* (version 2.0.0 Wickham et al., 2019), *semTools* (Version 0.5-10

Jorgensen et al., 2026) and *lavaan.mi* (version 0.1-1 Jorgensen, 2026). Data and annotated syntax of the analyses are assessable on a repository of the *Open Science Framework* https://osf.io/b2ugc/overview?view_only=64f208fd19ef48909404f2eb3ddaa496.

**Latent Modeling**

The models were computed using *Confirmatory Factor Analysis* (CFA). For continuous data, parameters were estimated via *Maximum Likelihood with robust standard errors (MLR)*. For dichotomous the data, the *Weighted Least Squares Means and Variance-adjusted (WLSMV)* estimator was applied (Beauducel & Herzberg, 2006; Muthén et al., 1997). All latent models were identified by fixing the latent variance of each factor to one.

A model was evaluated as good fit if Scaled $CFI \geq .95$ and Scaled $RMSEA \leq .06$ (Hu & Bentler, 1999). For acceptable model fit the following boundaries were used: Scaled $CFI \geq .90$ and Scaled $RMSEA \leq .08$ (Bentler, 1990; Browne & Cudeck, 1992). McDonald's ω (McDonald, 1999) was used as an estimate of reliability.

0.2% of the data was missing. Missing data resulted from issues with technical devices during MTCI assessment, making it Missing Completely at Random (Heitjan & Basu, 1996). In models on item-level, missing values (n = 3) were handled by *full information maximum likelihood* estimation (Enders, 2022). When constructing parcels, missing values at the item level were excluded from the parcel mean calculation only for the respective individual. Cases with missing data at the test level (n = 1) were excluded only from analyses involving the MTCI.

**Parceling**

Due to the large number of indicators and the limited sample size, we opted for item parceling, in which multiple items of the same scale are averaged into composite scores (Little et al., 2002). The analysis was guided by the report standards of Sterba and Rights (2023). Since the tests did not contain subscales, there was no clear theoretical basis for assigning items to specific parcels. Therefore, we randomly assigned items to parcels within each construct. Parcels compromised between 3 and 4 items.

Unidimensionality of the constructs is considered a prerequisite for parceling (Little et al., 2002; Marsh et al., 2013; Sterba & Rights, 2023). To assess this assumption, we examined the fit of a g-factor model using confirmatory factor analysis (CFA) at the item level for each construct (see appendix). We decided to exclude those with a negative part-whole correlation ($n$ = 1), rather than excluding all items that contributed to poorer model fit. This approach was chosen to retain the breadth of measurement. Results using items fitting a unidimensional model can be found in the appendix. Please note that the pattern of results as well as the significance of parameters did not change.

Random allocation is arbitrary, and different allocations can lead to varying results, even when the construct is unidimensional (Sterba, 2019). This variability, known as Parcel Allocation Variability (PAV), should be accounted for in the models (Sterba & MacCallum, 2010). To address PAV, multiple item-to-parcel allocations were generated, and the results were pooled following Rubin’s rules(Rubin, 1987; Sterba & Rights, 2016). The number of random draws was determined using the *poolMalloc* function (available in semTools). In this approach, the number of repeated parcel allocations within the sample is progressively increased, stopping when the results stabilize and no longer show significant changes with additional allocations (Sterba & Rights, 2016). Additionally, we evaluated how much of the variability in parameter estimates can be specifically attributed to differences in parcel allocation. To do so, we calculated both the proportion of the total variance explained by PAV (PPAV) and the ratio of between-allocation variance to within-allocation variance (RPAV) for each estimate, the results can be found in the appendix. Following Sterba & Rights (2016), we initially started with 5 item-to-parcel allocations and increased the number of allocations by 5 models per iteration. A convergence criterion of 1% change in both pooled parameter estimates and pooled standard errors was applied.

# Results

## Descriptive Results

All gf-related measures showed a broad range of item difficulties and sufficient item discrimination. In figural BEFKI, items had a mean difficulty of $M$ = .68 (range: .26 - .97, please note the guessing probability of .11) and a mean item discrimination of .31. Set items showed a mean difficulty of $M$ = .49 (range: .17 to .91). The items had a mean discrimination of .40. The average item difficulty of ARC-AGI was $M$ = .60, with item difficulty ranging from .12 to .93 (see Figure 4). The mean item discrimination was .39. Overall, ARC-AGI demonstrated psychometrically sound characteristics on a descriptive level.

In MTCI, the mean creativity score across all items was $M$ = .47. The lowest creativity score for an item was .05, and the highest was .78. Creativity scores ranged widely for all items (.37 - .64). Item 2 had the greatest range in creativity scores (*min* = .05, *max* = .69, range = .64).

**Figure 4**

*Item Difficulties of ARC-AGI 1 Items in Humans*

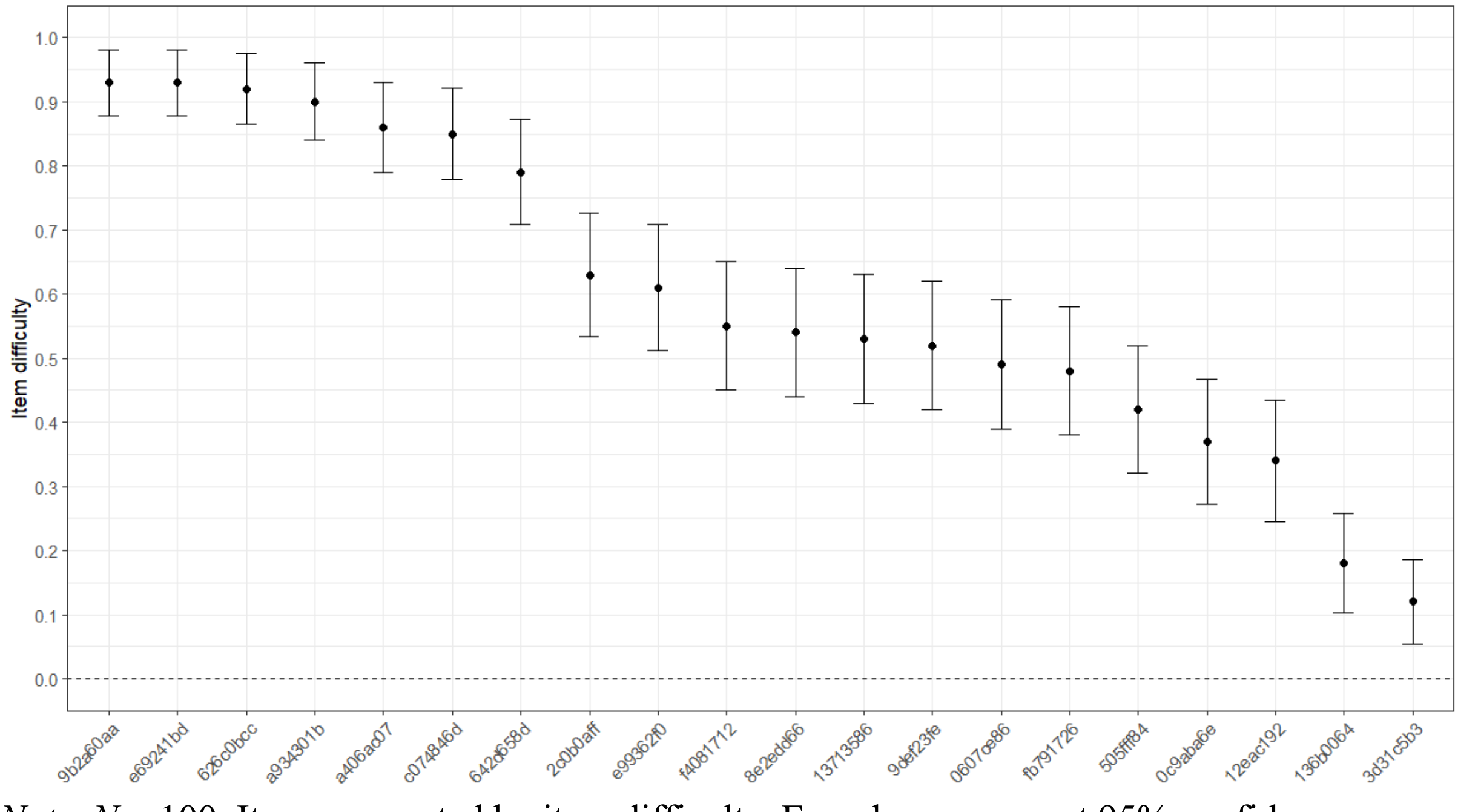

*Note.* $N$ = 100, Items are sorted by item difficulty. Error bars represent 95% confidence intervals.

Manifest correlations at the test level are displayed in Table 1. Figural BEFKI and ARC-AGI were correlated with $r = .41$ ($p < .001$), indicating a moderate effect. The correlation of Set and ARC was also moderate ($r = .33$, $p < .001$). ARC and MTCI showed a small correlation of $r = .10$ ($p = .153$). The correlation between Set and figural BEFKI ($r = .18$, $p = .041$) was similar to the correlation between Set and MTCI ($r = .17$, $p = .045$). Both were significant in a one-sided test.

**Table 1**

*Correlation Matrix on Test Level*

| | Figural BEFKI | Set | MTCI |
|---|---|---|---|
| Set | .18* | | |
| MTCI | -.05 | .17* | |
| ARC-AGI | .41* | .33* | .10 |

*Note.* pairwise *n* ranges from 99 to 100, * = significant to $p = .05$ in a one-sided test, figural BEFKI = figural subtest of Berlin Test of Fluid and Crystallized Intelligence, MTCI = Multi-Trial Creative Ideation task, ARC-AGI = Abstraction and Reasoning Corpus of Artificial General Intelligence version 1.

## Confirmatory Factor Analyses (CFA)

### *Measurement Model ARC-AGI*

The g-factor model for the ARC-AGI demonstrated a good fit ($X^2(5) = 6.60$, ($SD$ = 3.60), $CFI = .976$ ($SD = .027$), $RMSEA = .047$ ($SD = .044$)). All parcel indicators showed substantial and significant loadings. The reliability of the ARC-AGI factor was sufficient to good, with $\omega = .74$ ($SD = .015$). A total of $M = 35$ item-to-parcel allocations were used. In summary, these results showed that a g-factor model is fitting the ARC-AGI human data well.

### *Figural BEFKI and ARC-AGI*

The correlated factor model of figural BEFKI and ARC-AGI yielded $X^2(34) = 42.40$, ($SD = 12.20$) and demonstrated acceptable to good fit indices with $CFI = .943$ ($SD = .055$) and $RMSEA = .042$ ($SD = .032$). The two factors were substantially correlated ($\rho = .62$, 95%

CI [.38, .86]; see Figure 2), which is first evidence for the convergent validity of ARC-AGI in humans. Results are based on *M* = 55 item-to-parcel allocations.

**Figure 5**

*Correlated Factor Model of ARC-AGI and Figural BEFKI*

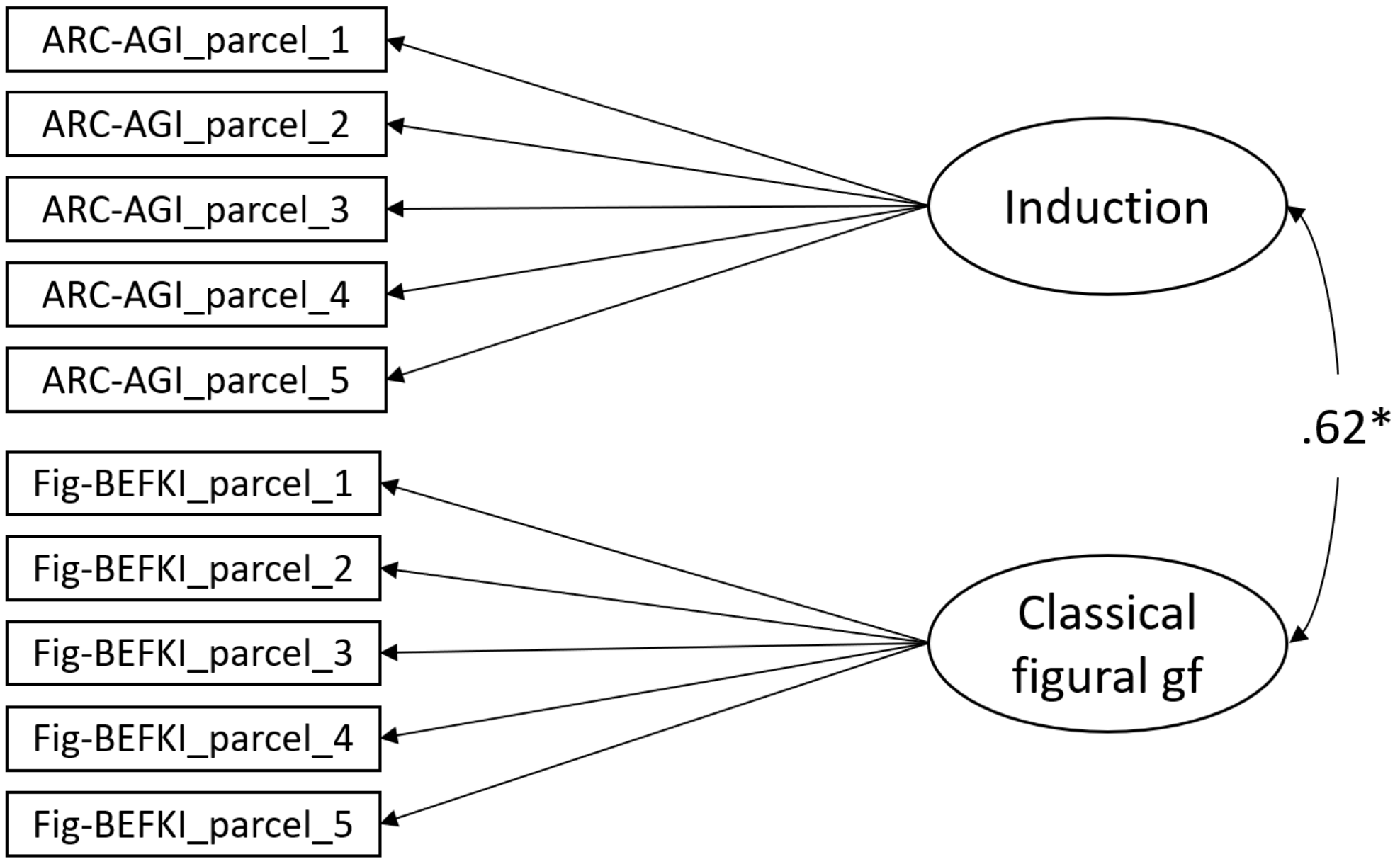

*Note.* *N* = 100; Number of parcel allocations = 55; ARC-AGI = Abstraction and Reasoning Corpus of Artificial General Intelligence version 1, figural BEFKI = figural subtest of Berlin Test of Fluid and Crystallized Intelligence, * = significant to *p* = .05.

***Full Correlation Model***

The $X^2$ value of the pooled CFA was $X^2(73) = 88.80$ (*SD* = 13.70). Model fit was acceptable to good, with *CFI* = .940 (*SD* = .044) and of *RMSEA* = .041 (*SD* = .023). ARC-AGI and Set correlated at $\rho = .49$ (95% CI [.22, .76]; see Figure 3), also supporting the convergent validity of ARC-AGI. The figural BEFKI and Set showed a correlation of $\rho = .27$ (95% CI [-.05, .59]). MTCI correlated at .15 with ARC-AGI (95% CI [-.14, .44]), at .25 with Set (95% CI [-.05, .54]) and at -.08 with figural BEFKI (95% CI [-.38, .23]). Reliabilities were: $\omega_{ARC-AGI} = .73$ (*SD* = .02), $\omega_{figural\ BEFKI} = .62$ (*SD* = .03), $\omega_{Set} = .61$ (*SD* = .07) and $\omega_{MTCI} = .80$ (*SD* = .02). *M* = 120 item-to-parcel allocations were used.

**Figure 6**

*Correlated Factor Model of ARC-AGI, figural BEFKI, SET and MTCI.*

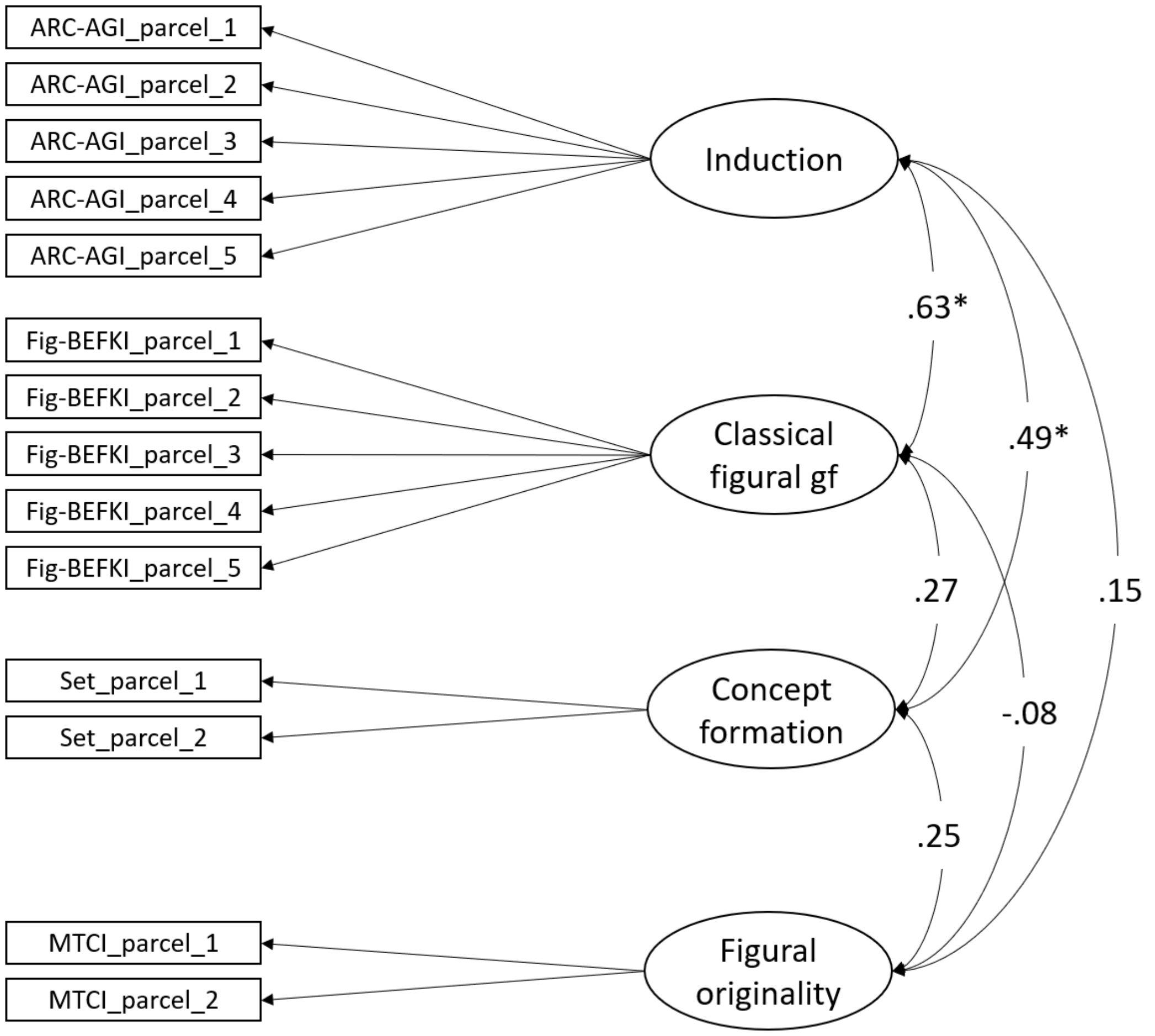

*Note.* $N = 99$; Number of parcel allocations = 120, ARC-AGI = Abstraction and Reasoning Corpus of Artificial General Intelligence version 1, figural BEFKI = figural subtest of Berlin Test of Fluid and Crystallized Intelligence, MTCI = Multi-Trial Creative Ideation task, * = significant to $p = .05$.

## Discussion

The present study had two aims. First, we provided an initial psychometric evaluation of ARC-AGI in a human sample. Second, we examined the extent to which ARC-AGI relates to established measures of fluid intelligence and figural originality in order to investigate its nomological net. ARC-AGI demonstrated good psychometric properties and showed a substantial latent correlation with figural BEFKI ($\rho = .63$), as well as a moderate association with concept formation. Relationships with figural originality of all tasks were weak, with the strongest effect observed between Set and MTCI ($\rho = .25$).

### ARC-AGI as Induction

The present study provides evidence for the usability of ARC-AGI in human ability measurement, as item difficulty varied, a g-factor model provided a good fit, and sufficient reliability was observed. Whereas previous studies collecting human performance data on ARC-AGI focused primarily on mean human performance (Johnson et al., 2021; LeGris et al., 2024), the present study provides the first evidence regarding its psychometric properties. Furthermore, a correlation to classical figural intelligence measures was shown. This finding was consistent with both perspectives on fluid intelligence considered in the present study: the view of fluid intelligence as the ability to process novel information and the view of fluid intelligence as being strongly influenced by working memory capacity.

Importantly, the results support the idea that fluid intelligence tasks can be extended to include tasks based on a more diverse set of underlying rules. ARC-AGI is characterized by a wide variety of rules (Chollet, 2019), in contrast to many established measures of fluid intelligence, which offer limited novelty by relying on a restricted and recurring set of underlying rules (e.g., Becker et al., 2016; Carpenter et al., 1990; Schroeders & Walter, 2026). Notably, the importance of novel information processing is emphasized in many definitions of fluid intelligence (Gignac & Szodorai, 2024; W. J. Schneider & McGrew, 2018; Sternberg, 1981, 1984). In this regard, ARC-AGI may help broaden the operationalization of fluid intelligence and contribute to a more complete measure of the construct of fluid intelligence.

Following the perspective of fluid intelligence as novel information processing, the relationship between ARC-AGI and originality was examined. It was argued that the process of identifying underlying rules may be associated with the ability to generate unique, rare but valid ideas (i.e., originality). The correlations between all fluid intelligence measures and figural originality were small. Set and ARC-AGI showed descriptively higher associations with originality compared to figural BEFKI. Effect sizes were in the range typically reported for associations between fluid intelligence tasks and divergent thinking (Gerwig et al., 2021). These findings provide a first and preliminary indication that tasks with a stronger rule-induction component are more closely related with fluid intelligence than with divergent thinking. They do not imply that induction should be considered a component of divergent thinking.

This study should be viewed as an initial step toward a more systematic investigation of the role of rule discovery within fluid intelligence. Given the limited scope in terms of sample size and number of indicators, strong conclusions regarding the position of inductive rule discovery within the broader structure of cognitive abilities would be premature. Instead, we use the present findings to outline directions for future research and to evaluate the results in light of the two theoretical perspectives introduced earlier: fluid intelligence as novel information processing and fluid intelligence as working memory capacity. In the second part we discuss the meaning of the validation in humans in the context of AI benchmark validation.

**Three Models for Future Research**

The present findings are too preliminary to allow for adjudicating between competing theoretical accounts of fluid intelligence measures, but they enable to formulate three testable models.

*Model 1: Induction as determined primarily by working memory capacity*

The first model posits that all fluid intelligence tasks are interchangeable because they rely on working memory capacity. This is based on the high correlation typically found between fluid intelligence and WMC (Kyllonen & Christal, 1990), as well as findings indicating that the difficulty of fluid intelligence tasks is best predicted by working memory load (Carpenter et al., 1990; Cortes et al., 2021; Gabales & Birney, 2011; Hartung et al., 2022; Primi, 2001). According to this view, an induction measure such as ARC-AGI should exhibit high correlations with working memory capacity comparable in magnitude to those observed for classical gf tests. It should also display similar relationships as gf with other covariates, such as divergent thinking. If this were the case, the emphasis on novelty in definitional accounts of fluid intelligence would be dispensable, as the defining factor would not be the novelty of the task, but rather the working memory demands imposed by the task.

*Model 2: Induction as a component of divergent thinking*

In this model, induction is a key ingredient of divergent thinking. Induction should thus be distinguishable from both fluid intelligence and working memory capacity. At the same time, a stronger association with divergent thinking than to classical fluid intelligence tests should be demonstrated.

*Model 3: Induction as an underrepresented component of fluid intelligence*

This model reflects current theories of fluid intelligence, in which induction is considered a component of fluid intelligence (W. J. Schneider & McGrew, 2018). From this perspective, induction is already implicitly included in gf as it is theoretically defined, but as argued earlier, it is severely underrepresented in prototypical measurement, mostly due to a lack of psychometrically convincing tests. As Wilhelm (2005) found that gf tasks built content factors (i.e., figural/spatial, numerical, and verbal), an investigation of the relation of ARC-AGI to a more diverse set of gf-indicators would be relevant, to investigate whether this pattern also applies to this more induction focused test. Future research should additionally expand the set of rule-focused tasks to test whether they can be separated from typical fluid intelligence tasks, in contrast to earlier attempts to capture induction, such as insight problems (Chuderski & Jastrzębski, 2018; Sternberg, 1984), and thereby investigating whether rule focused tasks may capture variance not assessed by standard measures of fluid intelligence. It would be necessary to demonstrate a high correlation between different components of gf, including induction, but also to establish a clear distinction from working memory capacity and divergent thinking respectively. If induction is part of gf, similar relationships of induction and typical measures of gf to other constructs would be expected.

## What Makes a Good Induction Measure?

No matter which theoretical perspective is adopted, a more precise understanding of the role of rule induction requires the development and application of sound inductive measures. In this study, both ARC-AGI and Set are used as measures of induction, with Set classified as a measure of concept formation.

Although concept formation is commonly classified as induction (Carroll, 1993; W. J. Schneider & McGrew, 2018), assessments remain rare. In the current study, Set was used as an operationalization. However, its suitability as a measure of induction is debatable, as the grouping rules are explicitly provided in the instructions. However, the task requires participants to group stimuli based on shared or non-shared attributes which is part of concept formation (Bourne, 1966; Seel, 2012). Thus, Set captures only part of the concept formation process and lacks the rule-identification component emphasized by Hunt (1961).

A stronger operationalization of induction would more closely align with the view proposed by Adkins and Lyerly (1952, p. 65) that “once the rules are understood, any subject can solve them correctly given sufficient time“. Although formulated for concept formation, this can be applied to induction tasks more generally, as it highlights the distinction between rule discovery and rule application. Ideally, performance in an induction task should be determined entirely by the successful inference of rules. Accordingly, rules should not recur across tasks, as repetition may shift the emphasis from induction to application. ARC-AGI, involves a highly diverse set of rules and provides a good example of inductive tasks and also illustrates how intelligence research can benefit from AI benchmarks.

## Relevance of ARC-AGI Validation in AI-Evaluation

The results of our study indicate promising psychometric properties of the ARC-AGI benchmark in a larger sample. Consistent with previous studies, participants were able to solve the tasks, and item difficulty varied (Johnson et al., 2021; LeGris et al., 2025). In contrast to earlier studies, all participants completed the same 20 ARC-AGI items, ensuring a large number of observations per item. This design allowed us to examine the correlations between tasks. The substantial correlations with fluid intelligence measures provide initial evidence for the convergent validity of ARC-AGI as a measure of fluid intelligence. ARC-AGI has previously been proposed as a measure of human fluid intelligence (Chollet, 2019), but an empirical test of this claim was pending.

On top of these findings, the present study highlights an often underemphasized aspect of AI benchmark evaluation, namely the extent to which benchmarks correspond to the corresponding constructs in humans (Sühr et al., 2025). Benchmarks in AI evaluation have faced criticism regarding their psychometric properties, especially regarding lacking validity (Alaa et al., 2025; Bean et al., 2025; Binette & Reiter, 2024; Eriksson et al., 2025). In the meantime, many frameworks have been released in the field of computer science to guide a way to more valid measures (Bean et al., 2025; Binette & Reiter, 2024; Reuel et al., 2024; Salaudeen et al., 2025; Wallach et al., 2025), but none of them mentioned the link to analogous constructs or abilities in humans. In line with Sühr et al. (2025), we argue that AI benchmarks should be more systematically embedded within the nomological network of human cognitive abilities. Although such validation requires additional empirical effort, human data are essential for clarifying what benchmarks measure and whether they meaningfully relate to established constructs such as fluid intelligence. From this perspective, human-based validation can be considered an important initial step in establishing convergent validity for AI benchmarks. Once such relationships have been established for a given construct domain (e.g., fluid intelligence), benchmarks may serve as reference points for evaluating related tasks within the same domain in AI.

Finally, we emphasize that establishing empirical links between AI benchmarks and human cognitive abilities should not be interpreted as proof of intelligence in machines. In particular, such findings do not justify the inference that performance of machine systems on the same tasks reflects intelligence in a psychological sense, nor do they warrant the assumption that the same latent variable underlying human performance is present in artificial systems. Differences in underlying mechanisms preclude an equivalence between human latent constructs and machine performance, even if observable outcomes were identical.

**Further Directions**

In future studies, several directions should be considered. First, the proposed models of induction should be empirically tested with respect to their relationships with working memory capacity, divergent thinking, and classical measures of fluid intelligence. Such model comparisons are more convincing if additional assessment instruments of rule induction are included.

Second, further work should aim to integrate benchmark-based measures into the nomological network of human intelligence. Since the completion of this study, ARC-AGI 2 and ARC-AGI 3 have been released (ARC Prize Foundation, 2026; Chollet, Knoop, Kamradt, Landers, et al., 2025). ARC-AGI 2 extends the original format with more complex rules, while ARC-AGI 3 introduces an interactive, exploratory setting where users must infer task goals without examples. Whether these extensions assess the same underlying construct in humans remains an open empirical question. In addition, benchmarks targeting constructs beyond fluid intelligence should be integrated into the nomological network of human cognition. This may contribute to a clearer and more structured organization of proposed benchmarks and their corresponding cognitive constructs.

Third, future research should continue to explore how AI benchmarks can be used to extend intelligence assessment. Benchmarks developed within the AI community can provide novel measurement approaches, particularly when they capture aspects that are not captured with traditional measures.

**Limitations**

In this study, we examined ARC-AGI as a potential measure of inductive fluid intelligence, but several limitations should be acknowledged. Due to the relatively small and highly educated sample, parameter estimates are associated with substantial uncertainty. Consequently, these estimates may change in future, larger and more diverse samples. In addition, items were randomly combined into parcels to estimate latent models. While this approach made model estimation feasible, and the parceling strategy proposed by Sterba & Rights (2016) accounts for variance introduced through the creation of multiple parcels, it also entails drawbacks. Random parceling can obscure construct-relevant variance within parcels, thereby masking potential model misspecification and artificially inflating model fit. If underlying assumptions are violated, this may also bias associations with other constructs (Marsh et al., 2013).

All latent factors in this study consisted of one task. While this enabled us to assemble different constructs, the constructs are not represented in their full scope. For example, measures of fluid intelligence were based on figural content. Performance on a single task reflects not only the intended latent construct but also task-specific variance, such as stimulus characteristics (Schmiedek et al., 2014; Wilhelm et al., 2025). Accordingly, a broader and more heterogeneous set of indicators per construct would more adequately capture its core (Little et al., 1999).

**Conclusion**

The AI benchmark ARC-AGI represents a promising measure of human cognitive abilities. We found substantial correlations with established measures of fluid intelligence below unity. As ARC-AGI, in contrast to many established gf measures, covers a diverse set of rules, these findings motivate further investigation of tasks that emphasize novel information processing. Although novelty is a core component of many definitions of fluid intelligence, it is underrepresented in commonly used induction tasks. Consequently, the use of benchmarks in intelligence research can contribute to a more comprehensive assessment of human cognitive abilities. In turn, validating AI benchmarks in human samples can help structure and improve the evaluation of artificial intelligent systems.

## Appendix

**Table A1**

*Fit of Item-level g-Models.*

| Test | # items | $X^2$(df) | *CFI* | *RMSEA* | *SRMR* | *ω* |
|---|---|---|---|---|---|---|
| Figural BEFKI | 15[1] | 109.1 (90) | .756 | .046 | .171 | .665 |
| Fig.BEFKI_unidim | 12 | 61.05 (54) | .894 | .036 | .155 | .673 |
| ARC | 20 | 196.5 (170) | .853 | .040 | .174 | .775 |
| ARC_unidim | 19 | 158.7 (152) | .963 | .021 | .154 | .777 |
| Set | 8 | 24.8 (20) | .950 | .049 | .125 | .683 |
| MTCI | 6 | 6.8 (9) | 1.00 | .000 | .027 | .808 |

*Note. N* = 100, [1] = one item removed due to negative discrimination.

**Table A2**

*Fit of ARC-AGI g-Factor Model With Items Fitting a Unidimensional Model.*

| $X^2$(SD) | *df* | *CFI (SD)* | *RMSEA (SD)* | *ω* (SD) |
|---|---|---|---|---|
| 10.20 (4.66) | 9 | .978 (.032) | .033 (.037) | .741 (.014) |

*Note. N* = 100, Number of Parcel Allocations = 45.

**Table A3**

*Fit of ARC-AGI and Figural BEFKI Model With Items Fitting a Unidimensional Model.*

| $X^2$(SD) | *df* | *CFI (SD)* | *RMSEA (SD)* | $\omega_{ARC-AGI}$(SD) | $\omega_{fig-BEFKI}$(SD) | *ρ* |
|---|---|---|---|---|---|---|
| 40.50 (9.35) | 34 | .955 (.037) | .037 (.028) | .732 (.017) | .632 (.030) | .56* |

*Note.* * = significant to p = .05, *N* = 100, Number of Parcel Allocations = 65.

**Table A4**

*Fit of the Full Correlation Model With Items Fitting a Unidimensional Model.*

| $X^2$ (SD) | *df* | *CFI (SD)* | *RMSEA (SD)* | $\omega_{ARC-AGI}$ (SD) | $\omega_{fig-BEFKI}$ (SD) | $\omega_{SET}$ (SD) | $\omega_{MTCI}$ (SD) |
|---|---|---|---|---|---|---|---|
| 86.90 (12.20) | 73 | .946 (.041) | .039 (.022) | .731 (.018) | .622 (.036) | .607 (.072) | .805 (.024) |

*Note. N* = 99, Number of Parcel Allocations = 85.

**Figure A1**

*Full Correlation Model With Items Fitting a Unidimensional Model.*

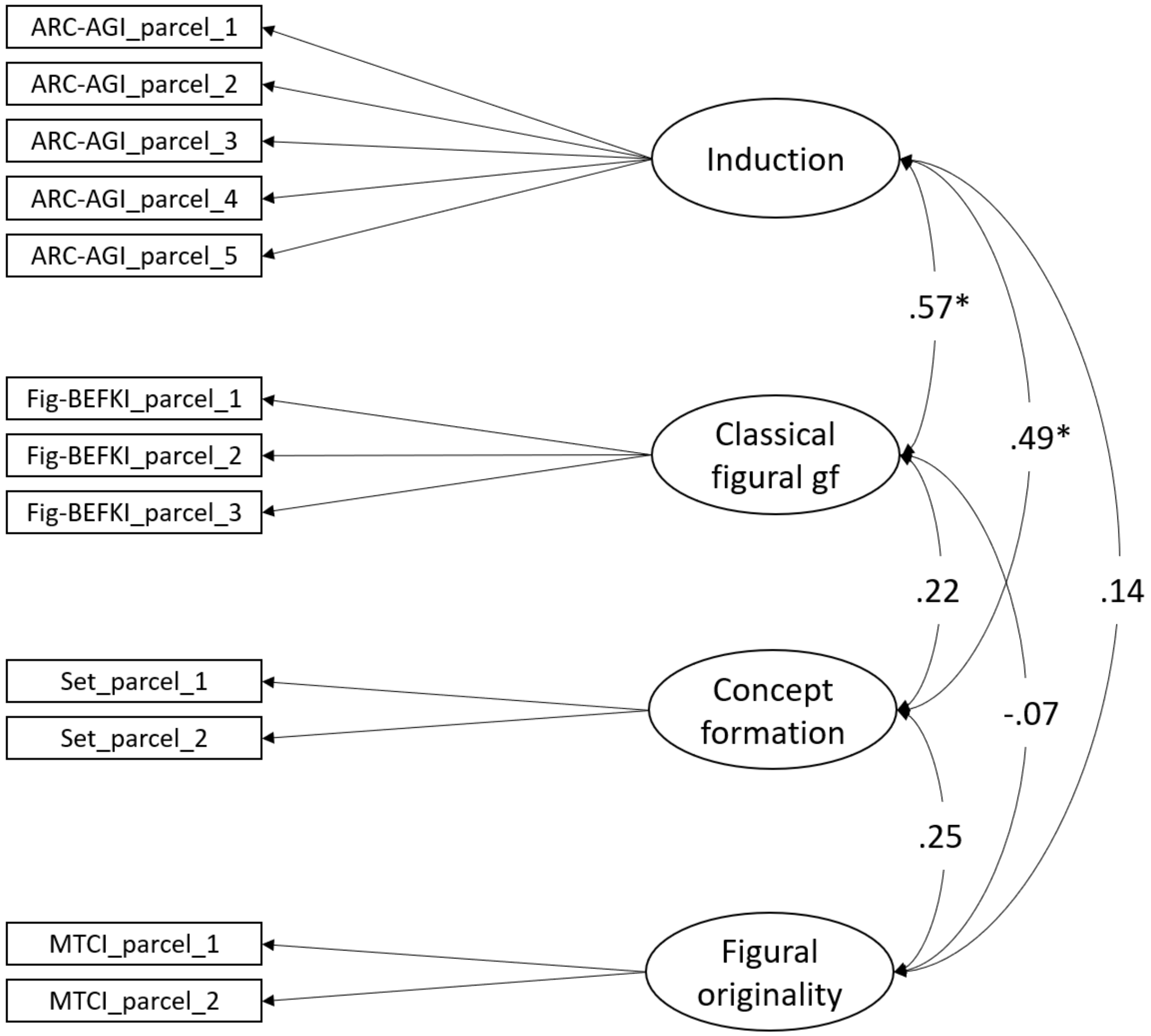

*Note. N* = 99.

**Table A5**

*Median PPAV and RPAV Values of Computed Models.*

| Model | *PPAV* | *RPAV* |
|---|---|---|
| ARC-AGI g -factor Model | 0.69 | 2.20 |
| ARC-AGI and figural BEFKI | 0.68 | 2.14 |
| Full correlation model | 0.68 | 2.10 |

*Note.* PPAV = Proportion of the total variance explained by Parcel Allocation Variance; RPAV = Ratio of between-Allocation variance to within-Allocation Variance.

**Table A6**

*Descriptive Statistics of ARC-AGI 1 in a Pilot Study.*

| Task Id | N | *Mean* | *SD* |
|---|---|---|---|
| d37a1ef5.json | 42 | .98 | .15 |
| 2546ccf6.json | 35 | .94 | .24 |
| 5783df64.json | 39 | .92 | .27 |
| a406ac07.json* | 34 | .88 | .33 |
| a934301b.json* | 30 | .87 | .35 |
| c074846d.json* | 29 | .86 | .35 |
| 60c09cac.json | 37 | .86 | .35 |
| 9b2a60aa.json* | 28 | .82 | .39 |
| 770cc55f.json | 37 | .81 | .40 |
| 1e97544e.json | 34 | .79 | .41 |
| e681b708.json | 27 | .78 | .42 |
| e69241bd.json* | 31 | .77 | .43 |
| 9772c176.json | 35 | .74 | .44 |
| 626c0bcc.json* | 27 | .74 | .45 |
| 2072aba6.json | 27 | .74 | .45 |
| 9a4bb226.json | 31 | .74 | .44 |
| 642d658d.json* | 39 | .74 | .44 |
| 93b4f4b3.json | 34 | .68 | .47 |
| e1baa8a4.json | 35 | .66 | .48 |
| 896d5239.json | 41 | .63 | .49 |
| 1d0a4b61.json | 38 | .61 | .50 |
| 13713586.json* | 128 | .59 | .49 |
| e9b4f6fc.json | 29 | .59 | .50 |
| c6e1b8da.json | 30 | .57 | .50 |
| 84db8fc4.json | 32 | .56 | .50 |
| f4081712.json* | 31 | .55 | .51 |
| bb52a14b.json | 38 | .55 | .50 |
| 0e671a1a.json | 37 | .54 | .51 |
| ed98d772.json | 34 | .53 | .51 |
| 2c0b0aff.json* | 30 | .53 | .51 |
| 15696249.json | 33 | .52 | .51 |
| e78887d1.json | 25 | .52 | .51 |
| 12422b43.json | 34 | .50 | .51 |
| 3490cc26.json | 32 | .50 | .51 |
| 0c9aba6e.json* | 26 | .50 | .51 |

| **Task Id** | N | ***Mean*** | ***SD*** |
|---|---|---|---|
| bf699163.json | 35 | .49 | .51 |
| d19f7514.json | 126 | .48 | .50 |
| 9def23fe.json* | 37 | .46 | .51 |
| 0607ce86.json* | 35 | .43 | .50 |
| 8e2edd66.json* | 34 | .38 | .49 |
| ac605cbb.json | 22 | .36 | .49 |
| 456873bc.json | 31 | .35 | .49 |
| e99362f0.json* | 30 | .33 | .48 |
| 96a8c0cd.json | 33 | .33 | .48 |
| 1acc24af.json | 29 | .31 | .47 |
| 7953d61e.json | 30 | .30 | .47 |
| 8719f442.json | 24 | .29 | .46 |
| fb791726.json* | 35 | .29 | .46 |
| 8fbca751.json | 32 | .28 | .46 |
| 85b81ff1.json | 27 | .26 | .45 |
| bc4146bd.json | 32 | .25 | .44 |
| 505fff84.json* | 25 | .24 | .44 |
| 50a16a69.json | 33 | .18 | .39 |
| 12eac192.json* | 28 | .18 | .39 |
| f8be4b64.json | 29 | .17 | .38 |
| 136b0064.json* | 39 | .15 | .37 |
| 0934a4d8.json | 32 | .12 | .34 |
| 3d31c5b3.json* | 32 | .03 | .18 |
| 08573cc6.json | 31 | .03 | .18 |

*Note.* Task Ids marked with * were used in the present study.